\documentclass[journal]{IEEEtran}

\usepackage{amssymb}
\usepackage{bbm}
\usepackage{mathbbol}
\usepackage{newtxmath}
\usepackage{amsmath}
\usepackage{graphicx}
\usepackage{subfigure}
\usepackage{caption}
\usepackage{algorithm}
\usepackage{algorithmic}
\usepackage[top=0.8in, bottom=0.8in, left=0.68in, right=0.68in]{geometry}
\begin{document}
%
\title{Continual Learning-Aided Super-Resolution Scheme for Channel Reconstruction and Generalization in OFDM Systems }

\author{Jianqiao Chen,
    Nan Ma,
    Wenkai Liu,
    Xiaodong Xu,
    and Ping Zhang
\thanks{
Jianqiao Chen is with the ZGC Institute of Ubiquitous-X Innovation and Applications, Beijing 100876, China. (e-mail: jqchen1988@163.com)
}
\thanks{Nan Ma, Wenkai Liu, Xiaodong Xu, and Ping Zhang are with the State Key Laboratory of Networking and Switching Technology, Beijing University of Posts and Telecommunications, Beijing 100876, China, and Xiaodong Xu is also with the ZGC Institute of Ubiquitous-X Innovation and Applications, Beijing 100876, China. (e-mail: manan@bupt.edu.cn, liuwenkai@bupt.edu.cn, xuxiaodong@bupt.edu.cn, pzhang@bupt.edu.cn)}
}

%


\maketitle
\thispagestyle{empty}
\begin{abstract}
Channel reconstruction and generalization capability are of equal importance for developing channel estimation schemes within deep learning (DL) framework.
In this paper, we exploit a novel DL-based scheme for efficient OFDM channel estimation where the neural networks for channel reconstruction and generalization are respectively designed.
For the former, we propose a dual-attention-aided super-resolution neural network (DA-SRNN) to map the channels at pilot positions to the whole time-frequency channels. 
Specifically, the channel-spatial attention mechanism is first introduced to sequentially infer attention maps along two separate dimensions corresponding to two types of underlying channel correlations, and then the lightweight SR module is developed for efficient channel reconstruction. 
For the latter, we introduce continual learning (CL)-aided training strategies to make the neural network adapt to different channel distributions. Specifically, the elastic weight consolidation (EWC) is introduced as the regularization term in regard to loss function of channel reconstruction, which can constrain the direction and space of updating the important weights of neural networks among different channel distributions. Meanwhile, the corresponding training process is provided in detail.
By evaluating under 3rd Generation Partnership Project (3GPP) channel models, numerical results verify the superiority of the proposed channel estimation scheme with significantly improved channel reconstruction and generalization performance over counterparts.
\end{abstract}


\IEEEpeerreviewmaketitle

\section{Introduction}

\IEEEPARstart{O}{wing} to some attractive features, such as the bandwidth efficiency and robustness to frequency selective fading channels, orthogonal frequency division multiplexing (OFDM) technology has been widely deployed in 5G and 5G-advanced systems [1], [2]. The promising gains brought by OFDM largely rely on acquiring accurate knowledge of channel state information (CSI), which is fundamental for subsequent communication processing, including the correlation detection, demodulation and equalization [3]. To cope with the fast-changing wireless environment, the pilot-aided channel estimation schemes for acquiring CSI is generally applicable. However, since the pilot overhead deteriorate the data rate, it puts pressure on designing efficient channel estimation schemes with the limited pilot resources. 



Deep learning (DL)-based approaches have demonstrated outstanding performance in various applications of physical layer communications [4], [5], which revolutionizes the forefronts of both
academia and industry. 
Focusing on the research field of channel estimation, the DL-based image super-resolution (SR) techniques [6] are creatively introduced [7]-[10]. 
In [7], the ChannelNet consisting of SR convolutional neural network (SRCNN) and denoising CNN (DnCNN) is proposed for both the channel interpolation and denoising. Moreover, the ChannelNet combined with concrete autoencoder is developed to obtain the most informative locations for pilots when estimating channels [8]. Although the SRCNN-based methods obtain better performance than the conventional methods, they have high-computational cost due to adopt the pre-upsampling procedure as to training data. Aided by residual neural network, a deep residual channel estimation network (ReEsNet) is designed and trained with the post-upsampling procedure, which therefore has high performance and low-computational cost [9]. To gradually accommodate large upsampling factors in channel estimation, a fully progressive image super-resolution scheme is proposed for dividing the entire estimation process into multiple stages, in which the LR image needs to be feature extracted and upsampled to a higher resolution [10]. Overall, these studies focus on developing efficient methods of feature extraction for improving channel estimation accuracy.  

On the other hand, the channel generalization capability is another very important aspect as to DL-based channel estimation methods for practical application. The aforementioned methods are trained and evaluated under the assumption of same channel distributions [7]-[10]. 
However, as to the inherently non-stationary propagation environment,  such as the different power delay profile (PDPs) and various transmission conditions (i.e., non-line-of-sight (NLoS) and line-of-sight (LoS)), the shift in channel distributions results in the degradation or even the failure of the trained neural networks in deployment phase.
To deal with the difficulty, the transfer learning technology is introduced to channel estimation procedure [11], [12]. In these cases, some layers of trained neural network are kept unchanged, while other layers are retrained for updating model parameters. So, the performance degradation due to channel mismatches can be alleviated with small amount of training data. Another similar approaches consider the initialization of the neural network as desired inductive bias by virtue of meta learning technology [13]-[15], which help accelerate convergence of neural network. Although these methods are efficient, they still consume computing resources for fine-tuning after deployment, which is difficult for some devices with limited computing power.

Considering different channel settings as channel estimation tasks, e.g., the signal-to-noise ratio (SNR) and coherence time, the continual learning (CL) technology is introduced to adapt to task changes by not forgetting the learned skill used in the previous task [16].
In this case, the well-trained neural network can be generalized to different channel settings without retraining.
However, it only tests several traditional CL algorithms with a fully-connected neural network, which are not specifically designed for pilot-aided channel estimation problems. 
Additionally, the impacts of different channel distributions on generalization capability of DL-based methods have not yet been studied.   
To fill up with these gaps, we propose a SR-inspired  neural network (SRNN) combined with CL-based training strategy for efficient channel estimation in OFDM systems.
Our contributions are outlined as follows.

\begin{itemize}
\item[]
\hspace{-0.5cm} $\bullet$
\hspace{-0cm}\textit{Channel Estimation Problem Formulation:} 
Considering both the channel reconstruction and generalization within DL framework, we newly formulate channel estimation as the SR problem under the condition of different channel distributions. 
Different from existing methods, it takes channel distribution as the condition when designing and training neural networks.

\hspace{-0.5cm} $\bullet$
\hspace{0.1cm}\textit{Efﬁcient Methods for Channel Estimation:} We design a novel SRNN combined with CL-based training strategies for channel reconstruction and generalization. 
For the former, the dual-attention modules are introduced for sequentially inferring attention maps, by which two different types of underlying channel correlations can be exploited for better feature extraction, which therefore results in improving channel reconstruction performance. 
For the latter, the elastic weight consolidation (EWC) [17] is introduced as the regularization term in regard to the loss function of channel reconstruction. By constraining the direction and space of updating the important weights of neural networks among different channel distributions, the channel generalization capacity can be improved.
\end{itemize}

\textit{Organization:} Section II builds the OFDM system model and formulates the corresponding channel estimation problem within DL framework.
Section III provides the proposed methods for efficient channel reconstruction and generalization.
Section IV presents numerical simulations for performance evaluation.
Section V concludes the paper.

\textit{Notation:} Boldface small letters denote vectors and boldface capital letters denote matrices. 
$\|\cdot\|_{F}$ denotes Frobenius norm.
$\mathcal{CN}(\cdot|\mu,\Gamma)$ denotes the complex Gaussian distribution with mean $\mu$ and covariance $\Gamma$.
$\mathbf{I}_{L}$ denotes an $L \times L$ identity matrix.
$\text{diag}(\cdot)$ denotes the diagonal operator.
$(\cdot)^{0^{-1}}$ denotes the Hadamard inverse (element wise inverse) operator.
$Re(\cdot)$ and $Im(\cdot)$ denote the acquisition of real part and imaginary part of a complex signal, respectively. 
$\circledast$ denotes the circular convolution operator.
$\odot$ denotes the dot product operator.



\section{System model and problem formulation}

\begin{figure}[t]
\centering
\includegraphics[width = 0.45\textwidth]{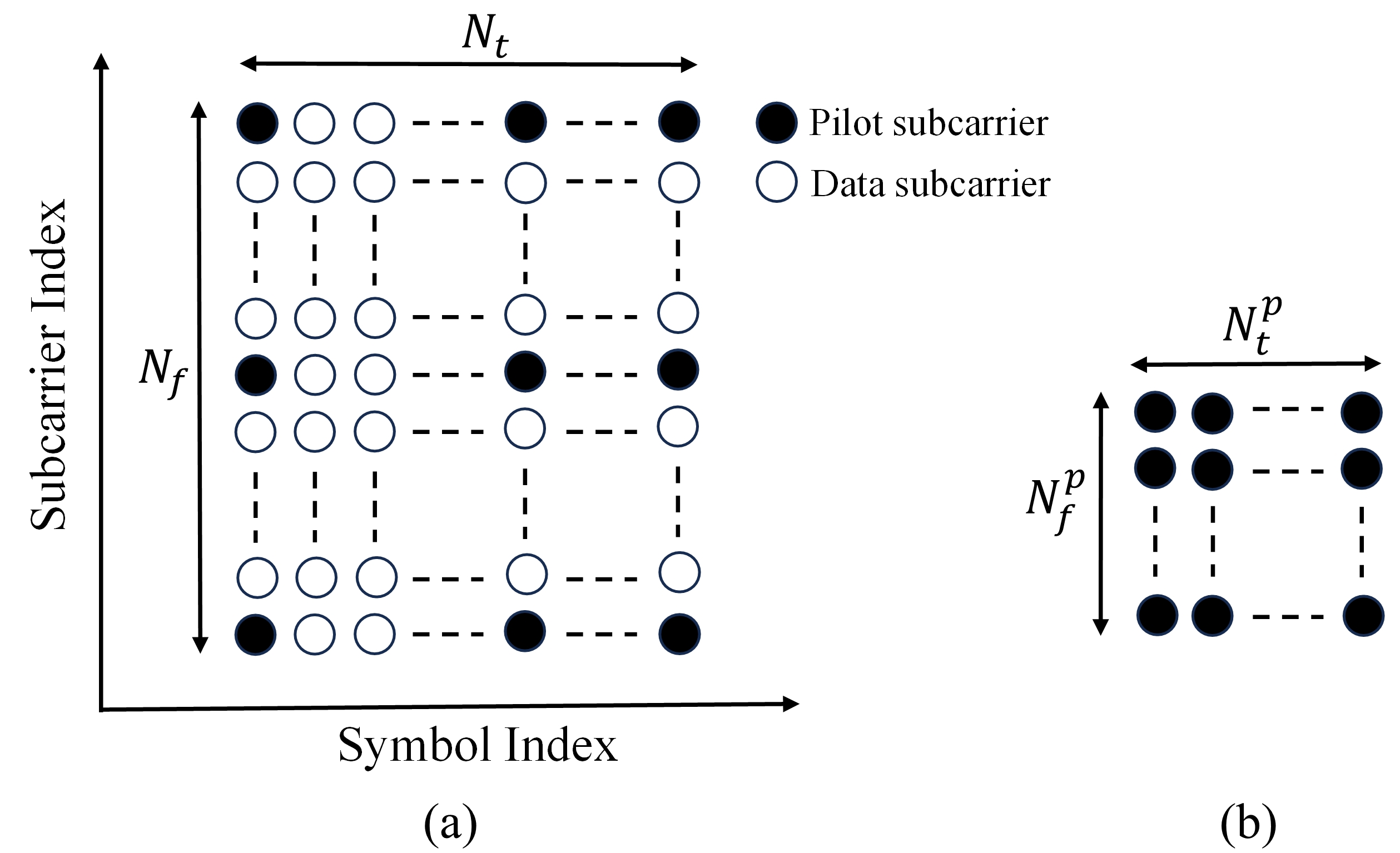}
\captionsetup{font={footnotesize}}
\caption{Illustration of time-frequency grids of OFDM subcarriers. (a) whole channel resources; (b) pilot resources.}
\label{fig1}
\end{figure}


This section commences with the formulation of the OFDM
channel estimation problem within DL framework, which considers both the channel reconstruction and generalization.
As shown in Fig. 1(a), we consider the channel estimation of an orthogonal frequency division multiplexing (OFDM) system, where the channel resources are divided into time-frequency grids of $N_{f}$ subcarriers and $N_{t}$ timeslots. 
The binary signals are first converted to the paralleled data stream with pilots inserted, which is then modulated on different subcarriers by inverse discrete-time Fourier transform (IDFT). To mitigate the inter-symbol interference, the cyclic prefic (CP) is embedded into OFDM symbols, whose length should be no shorter than the maximum delay spread of the channel. 
Assume $\textit{\textbf{s}}_{t} \in \mathbb{C}^{N_{f}\times 1}$ be the $t$th transmitted signal in time-domain, which can be transformed from the corresponding signals in frequency-domain by IDFT. The $t$th received signal in time-domain can be expressed as 
\begin{equation}
\textbf{\textit{y}}_{t} = \textbf{\textit{s}}_{t} \circledast \textbf{\textit{h}}_{t} + \textbf{\textit{w}}_{t}=\daleth_{s}
\begin{bmatrix} \textbf{\textit{h}}_{t}  \\ \textbf{\textit{0}}_{N_{f}-L} \end{bmatrix} + \textbf{\textit{w}}_{t},
\setcounter{equation}{1}
\label{eq1}
\end{equation}
where 
$\textbf{\textit{h}}_{t}\in \mathbb{C}^{L \times 1}$ denotes the time-domain CIRs with $L$ being the equivalent channel length, $\textbf{\textit{w}}_{t} \in \mathbb{C}^{N_{f} \times 1}$ denotes the additive Gaussian white noise (AWGN) with satisfying $\textbf{\textit{w}}_{t} \sim \mathcal{CN}(\textit{\textbf{w}}_{t}|0,\sigma^{2}\textbf{I}_{N_{f}})$, and $\daleth_{s} \in \mathbb{C}^{N_{f} \times N_{f}}$ denotes the Toeplitz matrix determined by $\textbf{\textit{s}}_{t}$, which can be expressed by using discrete Fourier transform (DFT) as 
\begin{equation}
\daleth_{s} = \textbf{F}_{N_{f}\times N_{f}}^{-1} \textbf{S}\textbf{F}_{N_{f}\times N_{f}},
\setcounter{equation}{2}
\label{eq2}
\end{equation}
where $\textbf{S}=\text{diag}(\textbf{\textit{s}}_{t})\in \mathbb{C}^{N_{f}\times N_{f}}$ and $\textbf{F}_{N_{f}\times N_{f}}$ denotes $N_{f}$-point DFT.

Then, the received signal in frequency-domain can be calculated as 
\begin{equation}
\textbf{\textit{y}}_{t}^{f}=\textbf{S}\textbf{F}_{N_{f}\times L}\textbf{\textit{h}}_{t} + \textbf{F}_{N_{f}\times N_{f}}\textbf{\textit{w}}_{t}, 
\setcounter{equation}{3}
\label{eq3}
\end{equation}
where $\textbf{F}_{N_{f}\times L}$ denotes the selection of the first $L$ columns of $\textbf{F}_{N_{f}\times N_{f}}$.

The received signals of all timeslots can be expressed as 
\begin{equation}
\textbf{Y}=\textbf{S}\textbf{H}+\textbf{W},
\setcounter{equation}{4}
\label{eq4}
\end{equation}
where $\textbf{Y}=[\textbf{\textit{y}}_{1}^{f}, \textbf{\textit{y}}_{2}^{f},...,\textbf{\textit{y}}_{N_{t}}^{f}]\in \mathbb{C}^{N_{f}\times N_{t}}$, 
$\textbf{H}=[\textbf{\textit{h}}_{1}^{f},\textbf{\textit{h}}_{2}^{f},...,\textbf{\textit{h}}_{N_{t}}^{f}]\in \mathbb{C}^{N_{f}\times N_{t}}$ with $\textbf{\textit{h}}_{t}^{f}=\textbf{F}_{N_{f} \times L}\textbf{\textit{h}}_{t}$, and $\textbf{W}=[\textbf{\textit{w}}_{1}^{f},\textbf{\textit{w}}_{2}^{f},...,\textbf{\textit{w}}_{N_{t}}^{f}]\in \mathbb{C}^{N_{f}\times N_{t}}$ with $\textbf{\textit{w}}_{t}^{f}=\textbf{F}_{N_{f} \times N_{f}}\textbf{\textit{w}}_{t}\in \mathbb{C}^{N_{f}\times N_{t}}$.

To improve spectral efficiency, the uniform grid-type pilot-aided channel estimation where the $N_{f}^{p}$ subcarriers and $N_{t}^{p}$ timeslots are used for pilot transmission, as shown in Fig. 1. Assume the set of subcarriers that carry pilots $\textbf{\textit{p}}_{f}=\{p_{1}^{f}, p_{2}^{f}, ..., p_{N_{f}^{p}}^{f}\}$ with $\textbf{\textit{p}}_{f} \subset \{1,2,...,N_{f}\}$, and the set of timeslots that carry pilots $\textbf{\textit{p}}_{t}=\{p_{1}^{t}, p_{2}^{t}, ..., p_{N_{t}^{p}}^{t}\}$ with $\textbf{\textit{p}}_{t} \subset \{1,2,...,N_{t}\}$. The received signals corresponding to $\textbf{\textit{p}}_{f}$ and $\textbf{\textit{p}}_{t}$ can be expressed as
\begin{equation}
\textbf{Y}_{p}=\textbf{S}_{p}\textbf{H}_{p}+\textbf{W}_{p},
\setcounter{equation}{5}
\label{eq5}
\end{equation}
where $\textbf{Y}_{p}$, $\textbf{S}_{p}$, $\textbf{H}_{p}$ and $\textbf{W}_{p}$ are the received signal, pilot signal, channel coefficient, and noise signal, respectively.

Our goal is to accurately recover the whole time-frequency channel $\textbf{H}$ based on $\textbf{Y}_{p}$ and $\textbf{S}_{p}$, which usually contains two steps. Specifically, the channels at pilot positions are first estimated, and then the channels at non-pilot positions are predicted. 
However, obtaining accurate $\textbf{H}$ is difficult due to the limited number of pilots and inefficient interpolation methods. 
To deal with it, some SRNNs are recently developed to map the channel coefficients at pilot positions to the corresponding whole time-frequency channel coefficients in an end-to-end manner.
However, there are still two key considerations remain.
Firstly, the correlation among channel coefficients in $\textbf{H}_{p}$ are not fully exploited for feature extraction when training neural networks. 
Secondly, the specific data-driven SRNN cannot generalize to different channel distributions, which greatly limit their application in practice. Considering these issues, we formulate the channel estimation problem as follows:
\begin{subequations}\label{eqn-4}
  \begin{align}
    & \mathop{argmin}\limits_{\hat{\textbf{H}}^{(d)}} \left\| \hat{\textbf{H}}^{(d)} - \textbf{H}^{(d)} \right\|_{F}^{2},  \\
    & \text{s.t.} \hspace{0.38cm}  \textbf{H}^{(d)}=\mathcal{F}_{SR}\left(\Theta; \hat{\textbf{H}}^{(d)}_{p} \right), \\
    &  \hspace{0.8cm} \hat{\textbf{H}}^{(d)}_{p}\in \left\{\hat{\textbf{H}}^{(1)}_{p}, \hat{\textbf{H}}^{(2)}_{p},...,\hat{\textbf{H}}^{(D)}_{p}   \right\},
  \end{align}
\end{subequations}
where $\textbf{H}^{(d)}$,  $\hat{\textbf{H}}^{(d)}$ and $ \hat{\textbf{H}}^{(d)}_{p}$ denote the true whole channel, the reconstructed whole channel and the estimated pilot channel under the $d$th channel distribution, respectively, $D$ denotes the number of channel distributions, and $\mathcal{F}_{SR}\left(\Theta;\cdot\right)$ denotes the designed SRNN with the parameter set $\Theta$, which considers the above two issues. Hence, the remaining problem is to design a channel estimation scheme that 
minimizes channel reconstruction error with considering generalization capability to different channel distributions.

\begin{figure}
\centering
\subfigure[]{
  \label{fig:subfig:a}
   \includegraphics[width = 0.45\textwidth]{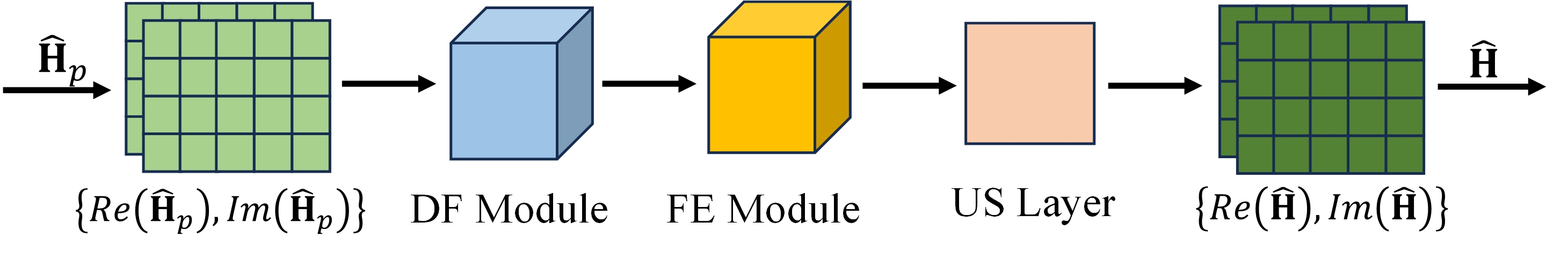}}
\hspace{1in}
\subfigure []{
 \label{fig:subfig:b}
 \includegraphics[width = 0.45\textwidth]{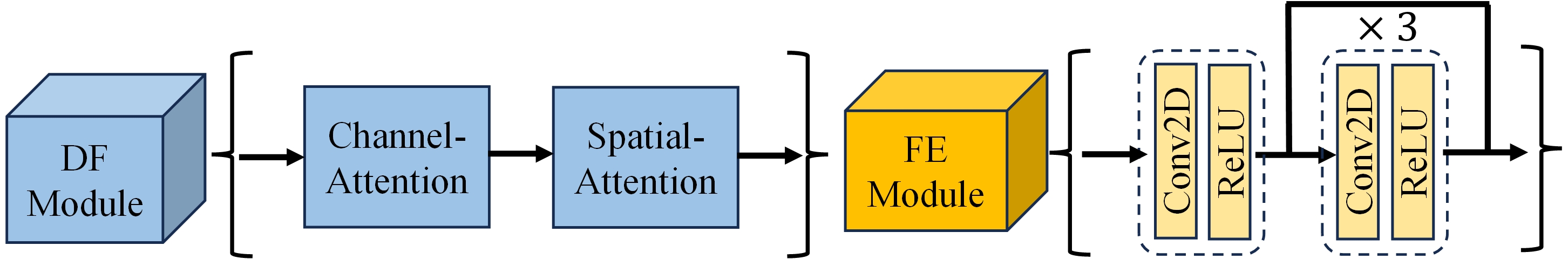}}
\hspace{1in}
\subfigure []{
 \label{fig:subfig:b}
 \includegraphics[width = 0.45\textwidth]{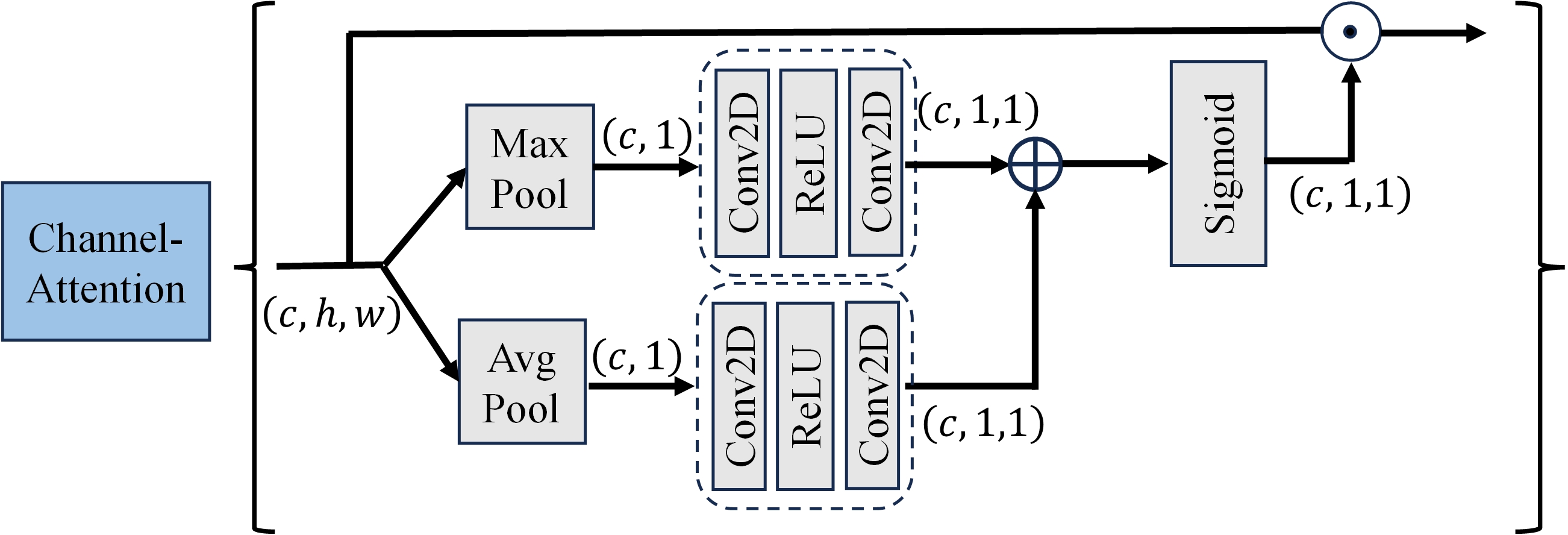}}
\hspace{1in}
\subfigure []{
 \label{fig:subfig:b}
 \includegraphics[width = 0.45\textwidth]{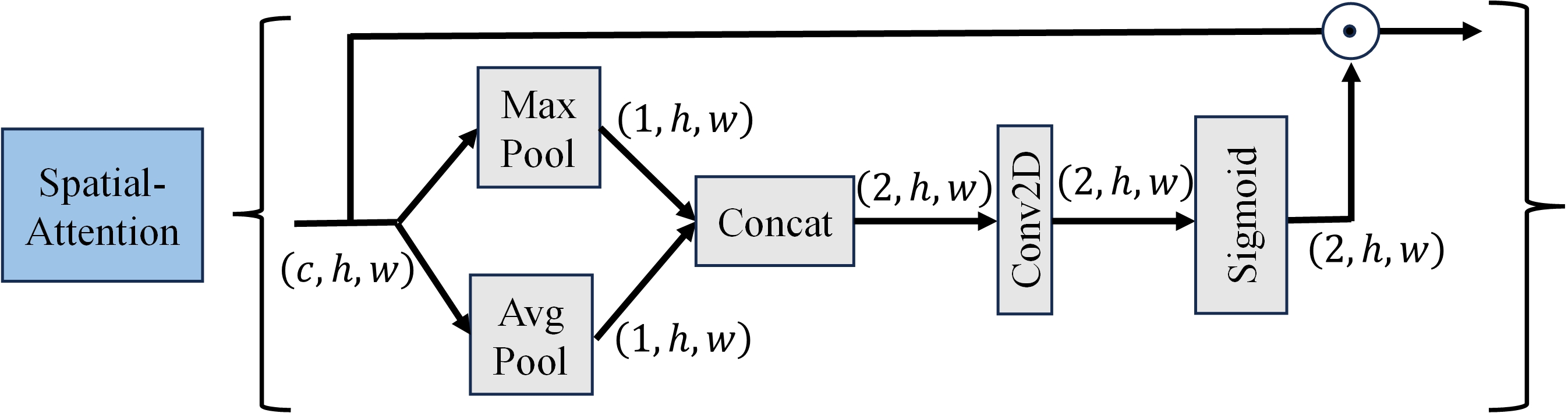}}
\captionsetup{font={footnotesize}}
\caption{Basic blocks of the dual-attention-aided super-resolution architectures. $(c, h, w)$ denotes the dimension of feature map with $c$ channel, $h$ height and $w$ width. Max Pool and Avg Pool denote the maximum pooling operation and average pooling operation, respectively. Concat denotes the matrix concatenation operation. Conv2D denotes the 2D convolution operation.}
\label{fig:subfig}
\end{figure}

\section{Channel Reconstruction and Generalization Mechanism}
\subsection{Dual-Attention-Aided SRNN for Channel Reconstruction}
The design of SRNN for channel reconstruction is provided in this section.
By using least square (LS) method [18], the estimated channels at pilot positions can be calculated as
\begin{equation}
\hat{\textbf{H}}_{p}=\textbf{Y}_{p} \odot \textbf{S}_{p}^{0^{-1}}.
\setcounter{equation}{7}
\label{eq7}
\end{equation}
Similar to image SR problems in computer vision (CV) field, we treat $\hat{\textbf{H}}_{p}$ as the training data to reconstruct the whole time-frequency channels $\hat{\textbf{H}}$ by mapping $\hat{\textbf{H}}_{p}$ to $\hat{\textbf{H}}$, namely $\hat{\textbf{H}}=\mathcal{F}_{SR}\left(\Theta; \hat{\textbf{H}}_{p} \right)$.
Note that the channel distribution identification is temporarily dropped for brevity. 

Since it is difficult for neural networks to deal with complex channel coefficients, $\hat{\textbf{H}}_{p}$ is turned into a set that consists of its real part and imaginary part for building the training data, namely $\left\{Re(\hat{\textbf{H}}_{p}), Im(\hat{\textbf{H}}_{p})\right\}$.
In this case, two different types of underlying channel correlations need to be exploited for efficiently training neural networks. Specifically, the first type is the correlation between the real part and imaginary part of channel coefficients, and the second type is the correlation among the channel coefficients within time-frequency resource block due to channel coherence.
Considering these two channel correlations, we design the dual-attention-aided SRNN (DA-SRNN), which consists of three parts, namely the data-fusion (DF) module, feature-extraction (FE) module and up-sampling (US) layer.
The architecture of DA-SRNN is illustrated in Fig. 2(a), and its
detailed descriptions are as follows.

\begin{figure}
\centering
\includegraphics[width = 0.45\textwidth]{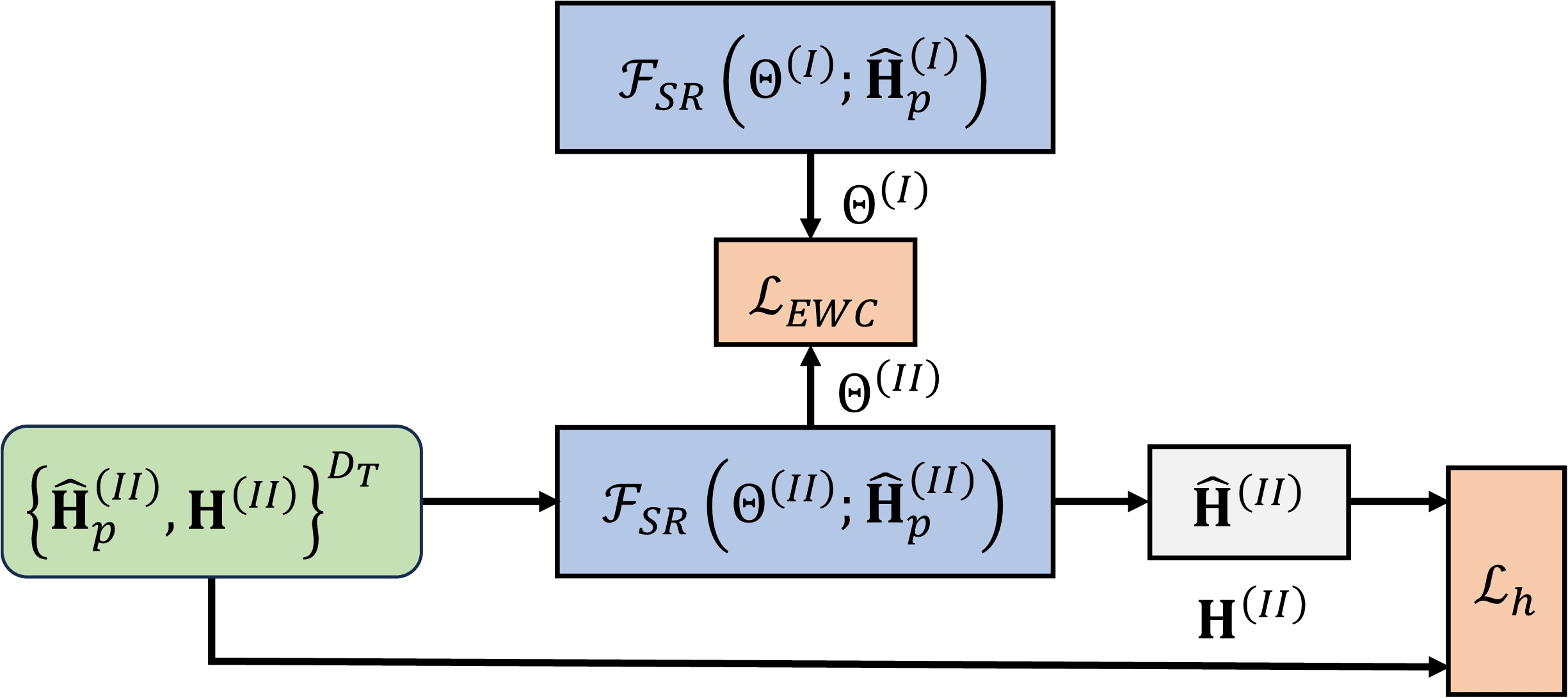}
\captionsetup{font={footnotesize}}
\caption{Illustration of structure diagram of CL-DA-SRNN scheme with considering two channel distributions.}
\label{fig3}
\end{figure}

For the DF module, the channel-attention and spatial-attention blocks are designed for sequentially inferring attention maps along two separate dimensions, which respectively correspond to two types of channel correlations.
To cope with high-dimensional time-frequency channel reconstruction, the convolution layers instead of fully-connected layers are adopted.
The detailed constituent parts are shown in Fig. 2(c) and Fig. 2(d).
For the channel-attention, the two-dimensional (2D) convolutional layer with 16 filters of size $3\times 3 \times 2$, which maps the input $N_{f}^{p} \times N_{t}^{p} \times 2$ to output of size $N_{f}^{p} \times N_{t}^{p} \times 16$. And then the following 2D convolutional layer has 2 ﬁlters of size $3 \times 3 \times 16$. For the spatial-attention, the 2D convolutional layer with 1 filter of size $7 \times 7 \times 2$.

The FE module is designed for feature extraction with a residual link. 
As shown in the right part of Fig. 2(b), the FE module consists of several 2D convolution and ReLU layers. The first 2D convolutional layer with 32 ﬁlters of size $5 \times 5 \times 2$ maps the input $N_{t}^{p} \times N_{t}^{p} \times 2$ to output of size $N_{t}^{p} \times N_{t}^{p} \times 32$. The second 2D convolutional layer with 16 ﬁlters of size $1\times 1 \times 32$, the third 2D convolutional layer with 16 ﬁlters of size $3\times 3 \times 12$, and the last 2D convolutional layer with 32 ﬁlters of size $1 \times 1 \times 16$. 
Combined with the DF module, it is found that the commendable feature extraction performance can be achieved by using shallow neural network.

Finally, the US layer adopts 2D deconvolution operation to map the feature to the whole time-frequency channel. Note that the post-upsampling procedure has the benefit of reducing cost of training the neural network.
The loss function for channel reconstruction is defined as 
\begin{equation}
\mathcal{L}_{h} = \frac{1}{| D_{s} |}\sum_{B_{s}\in D_{s} }\frac{1}{|B_{s}|}\sum_{\textbf{H}^{(B_{s})}\in \mathbb{C}^{B_{s}\times N_{f} \times N_{t} }} \left\| \hat{\textbf{H}}^{(B_{s})} - \textbf{H}^{(B_{s})} \right\|_{F}^{2},
\setcounter{equation}{8}
\label{eq8}
\end{equation}
where $|D_{s}|$ denotes the dimension of data set $D_{s}$, $B_{s}$ denotes the batch size of training sample, and $\hat{\textbf{H}}^{(B_{s})}$ denotes the reconstructed whole channel corresponding to true channel $\textbf{H}^{(B_{s})}$ with the batch size $B_{s}$. In training phase, the parameters of neural networks are dynamically adjusted by the backpropagation algorithm, and the near-global optimal solution is gradually approached by several iterations.

\begin{algorithm}
\begin{normalsize}
    \renewcommand{\algorithmicrequire}{\textbf{Input:}}
    \renewcommand\algorithmicensure {\textbf{Output:} }
    \captionsetup{font={normalsize}}
    \caption{CL-DA-SRNN Scheme}
        \textbf{Input:} $\left\{\hat{\textbf{H}}_{p}^{(I)}, \textbf{H}^{(I)} \right\}^{D_{T}}$, $\left\{\hat{\textbf{H}}_{p}^{(II)}, \textbf{H}^{(II)} \right\}^{D_{T}}$, $D_{T}$ denotes the data size for training. \\
        1: \hspace{0.1cm} randomly initialize the DA-SRNN parameters \\
        2: \hspace{0.1cm} consider $\left\{\hat{\textbf{H}}_{p}^{(I)}, \textbf{H}^{(I)} \right\}^{D_{T}}$ as the training data of task $I$, and train the DA-SRNN with (8) \\
        3: \hspace{0.1cm} calculate the FIM in regard to task $I$ by (10) \\
        4: \hspace{0.1cm} consider $\left\{\hat{\textbf{H}}_{p}^{(II)}, \textbf{H}^{(II)} \right\}^{D_{T}}$ as the training data of task $II$, and train the CL-DA-SRNN with (11)\\
        \textbf{Output:} CL-DA-SRNN parameters $\Theta_{CL-DA-SRNN}$
\end{normalsize}
\end{algorithm}

\subsection{CL-Aided Channel Generalization}
So far, the DA-SRNN has been designed for channel reconstruction. Next, we will treat it as the backbone network to further develop the method of channel generalization in regard to different channel distributions. 
We consider the different channel distributions as different tasks, whose channel data arrive sequentially for training.
In this case, the CL-aided training strategies are tailored to make the neural network adapt to sequentially changing tasks without forgetting the previously learned ones.
Specifically, the EWC is  introduced as the regularization term as to the loss function (8), which can constrain the direction and space of updating the important weights of neural networks among different channel distributions.

The proposed structure diagram is shown in Fig. 3, which takes two channel distributions as example.
Note that it can be extended to the scenario consisting of more channel distributions.
Without loss of generality, we assume $\hat{\textbf{H}}_{p}^{(I)}$ as the previous task and $\hat{\textbf{H}}_{p}^{(II)}$ as the latter task. The data of tasks sequentially arrive for training the neural network. As to latter task, the EWC term constrains the important weights to stay in low-error region for previous task, which is obtained by using a quadratic penalty combined with the diagonal approximation of fisher information matrix (FIM). The EWC loss function can calculated as 
\begin{equation}
\mathcal{L}_{EWC} = \sum_{i}\frac{\lambda}{2}F_{i}^{(I)}\left(\Theta_{i}^{(II)} - \Theta_{i}^{(I)}\right)^{2},
\setcounter{equation}{9}
\label{eq9}
\end{equation}
where $\lambda$ is the hyperparameter indicating how important the previous task is compared to the latter task, $i$ labels each parameter of neural network, and $F_{i}^{(I)}$ denotes the $i$th parameter corresponding to the task $I$ of FIM, which is approximately calculated as the average value of the gradient square as follows
\begin{equation}
F_{i}^{(I)} = \frac{1}{|D_{s}|}\sum_{B_{s} \in D_{s}}\frac{\partial^{2}\mathcal{L}_{h}\left(\Theta_{i}^{(I)}    \right)}{\partial\left(\Theta_{i}^{(I)}\right)^{2}}.
\setcounter{equation}{10}
\label{eq10}
\end{equation}

Finally, the total loss function can be expressed as 
\begin{equation}
\mathcal{L}_{total} = \mathcal{L}_{h} + \alpha\cdot\mathcal{L}_{EWC}, 
\setcounter{equation}{11}
\label{eq11}
\end{equation}
where $\alpha$ denotes the hyperparameters.
For clarity, we summarize our proposed scheme of considering both the channel reconstruction and generalization in \textbf{Algorithm 1}, which is referred as to the CL-aided DA-SRNN (CL-DA-SRNN) scheme.

\section{Performance evaluation}


This section presents numerical results to evaluate the performance of the proposed DA-SRNN and CL-DA-SRNN schemes, where the normalized mean square error (NMSE) is chosen as performance metrics. 

\subsection{Parameters Setup and Performance Metrics}
The parameters of the OFDM system are set as follows: carrier frequency $f_{c}=2$ GHz, the number of subcarriers $N_{f}=128$, the number of timeslots $N_{t}=28$, and quadrature phase shift keying (QPSK) modulation. The uniform grid-type pilot pattern is adopted, where the pilot intervals along the subcarrier dimension and timeslot dimension are set to be $9$ and $5$, respectively. The standard channel models in 3GPP TR 38.901 [19] is adopted to generate the channel coefﬁcients for training neural networks. Specifically, the tapped delay line (TDL) models, namely the TDL-A model and TDL-D model, are chosen. The TDL-A model only consists of NLoS multipaths, whereas the TDL-D model consists of LoS and NLoS multipaths. For more information of channel parameters, please refer to the Table 7.7.2-1 and Table 7.7.2-4 in [19]. Since they have different channel distributions, including the PDP and transmission conditions, it challenges the neural network design for channel reconstruction and generalization.

\begin{figure}
\centering
\subfigure[]{
  \label{fig:subfig:a}
   \includegraphics[width = 0.44\textwidth]{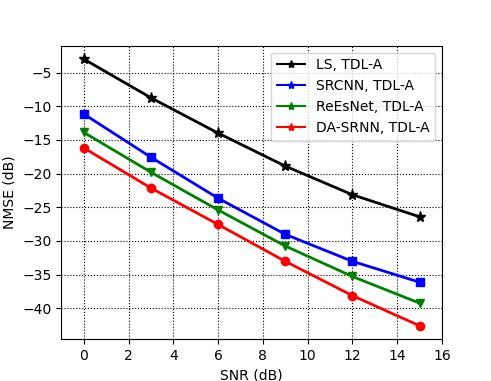}}
\hspace{1in}
\subfigure []{
 \label{fig:subfig:b}
 \includegraphics[width = 0.44\textwidth]{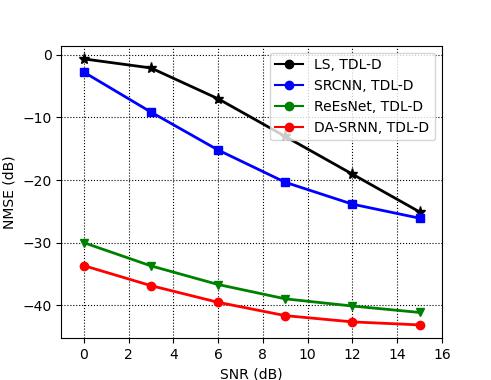}}
\captionsetup{font={footnotesize}}
\caption{NMSE of channel reconstruction versus SNR. (a) TDL-A model; (b) TDL-D model.}
\label{fig:subfig}
\end{figure}

Numerical simulations are conducted to investigate the performance of the proposed schemes, i.e., the DA-SRNN scheme and CL-DA-SRNN scheme, as well as compare with other related popular schemes, including the LS scheme [18], SRCNN scheme [7], ReEsNet scheme [9], and Multi-Task scheme. The LS scheme is the traditional method that combines with 2D linear interpolation. The SRCNN and ReEsNet schemes are most relevant schemes for channel reconstruction developed within DL framework. 
Note that the SRCNN and ReEsNet schemes are adopted to compare the channel reconstruction performance with the DA-SRNN scheme.
The multi-task scheme is adopted to train the DA-SRNN scheme with the mixed channel data of TDL-A and TDL-D models, which considered as the upper limit compared with the CL-DA-SRNN scheme. The NMSE is chosen as performance metrics, which can be calculated as
\begin{equation}
NMSE=\frac{1}{M_{c}}\sum_{m_{c}=1}^{M_{c}}\frac{\left\| 
\textbf{H}^{(m_{c})} - \hat{\textbf{H}}^{(m_{c})}  \right\|^{2}_{F}   }{\left\|\textbf{H}^{(m_{c})}\right\|^{2}_{F}},
\setcounter{equation}{12}
\label{eq12}
\end{equation}
where $\hat{\textbf{H}}^{(m_{c})}$ denotes the estimate of true channel matrix $\textbf{H}^{(m_{c})}$ at the $m_{c}$th Monte Carlo trail and $M_{c}$ is the total number of Monte Carlo trails. Overall, the $20000$ channel realizations with OFDM system configuration are created, among which the training samples contain $16000$ channel realizations, validation samples contain $1000$ channel realizations, and test samples contain $1000$ channel realizations. Clearly, the $M_{c}$ equals to $2000$ for performance evaluation. Mini-batch size of $128$, total epoch of $100$, and Adam optimizer with learning rate $0.001$ are adopted for the model training. 

\subsection{Channel Reconstruction Performance}
We first verify the capability of the DA-SRNN scheme for channel reconstruction performance, where the NMSEs of channel estimation versus SNRs are presented in Fig. 4. All the DL-based schemes are first trained by setting SNR=10dB, and then evaluated by SNR vector [0, 3, 6, 9, 12, 15]dB. It can be observed that: 1) NMSEs of all the schemes decrease as SNR increases, and the LS scheme obtains the worst performance. 2) Compared with SRCNN scheme and ReEsNet scheme, the DA-SRNN scheme obtains better NMSE performance due to exploiting the channel correlations by introducing dual-attention blocks.
3) Seeing from the comparison between Fig. 4(a) and Fig. 4(b), the LS scheme obtains the similar NMSE performance, whereas the other DL-based schemes show different NMSE performance trends.
So, it illustrates that specific data-driven DL-based schemes have different generalization capability in regard to different channel distributions. 

\begin{figure}[t]
\centering
\includegraphics[width = 0.44\textwidth]{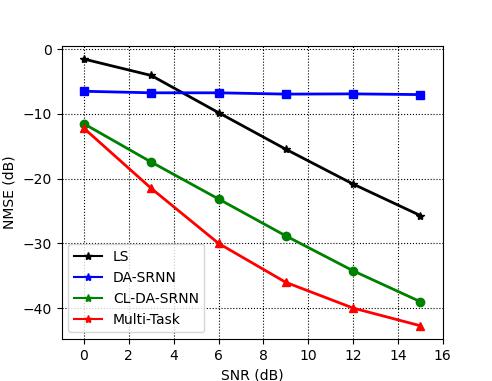}
\captionsetup{font={footnotesize}}
\caption{NMSE of channel generalization versus SNR.}
\label{fig1}
\end{figure}

\subsection{Channel Generalization Performance}
We verify the capability of the DL-DA-SRNN scheme for channel generalization performance, 
where the NMSEs of
channel estimation versus SNRs are presented in Fig. 5.
For the DA-SRNN scheme, it trained under the channel data of TDL-D model.
For the Multi-Task scheme, it trained under the mixed channel data of TDL-A and TDL-D models.
For evaluation, the channel data of TDL-A model and TDL-D model are mixed in order to imitate a practical scenario where different channel distributions exist simultaneously.
It can be observed that: 1) The DA-SRNN scheme shows the catastrophic forgetting phenomenon that the neural network trained with the TDL-D model cannot efficiently apply to the mixed channel data.
2) The Multi-Task scheme obtains the best NMSE performance. However, it requires larger training data sets and more training overhead. Moreover, it is faced with the problem of retraining the whole neural network as to additional channel data.
3) By introducing the CL training strategy, it shows that the CL-DA-SRNN achieves good generalization of channel reconstruction for different channel distributions. It is flexible that it only needs to be retrained on sequentially arrived channel data.

\section{Conclusion}
The problem of joint channel reconstruction and generalization for OFDM systems is tackled within DL framework. Specifically, the CL-DA-SRNN scheme is proposed for minimizing channel reconstruction error with considering generalization capability to different channel distributions.
On the one hand, the DA-SRNN is developed to map the channels at pilot positions to the whole time-frequency channels, where the dual-attention blocks are designed for efficient feature extraction as to underlying channel correlations.
On the other hand, the CL-aided training strategy is tailored to improve the channel generalization capability, where the EWC is introduced as the regularization term in regard to the reconstruction loss function. Meanwhile, the corresponding training process is provided in detail. 
By evaluating with 3GPP standard channel models, numerical results demonstrate that our proposed schemes attain superior performance to comparison schemes in terms of both channel reconstruction and generalization.


%




\ifCLASSOPTIONcaptionsoff
  \newpage
\fi

\end{document}